# Picture Collage with Genetic Algorithm and Stereo vision

Hesam Ekhtiyar[1], Mahdi Sheida[2] and Mahmood Amintoosi[3]

[1] Faculty of Electrical and Computer Engineering, Sabzevar Tarbiat Moallem University,
Sabzevar, Iran,
*hekhtiyar@gmail.com*

[2] Faculty of Electrical and Computer Engineering, Sabzevar Tarbiat Moallem University,
Sabzevar, Iran,
*m.sheida87@gmail.com*

[3] Faculty of Mathematics and Computer Science, Sabzevar Tarbiat Moallem University,
Sabzevar, Iran,
*amintoosi@sttu.ac.ir*

**Abstract**
In this paper, a salient region extraction method for creating picture collage based on stereo vision is proposed. Picture collage is a kind of visual image summary to arrange all input images on a given canvas, allowing overlay, to maximize visible visual information. The salient regions of each image are firstly extracted and represented as a depth map. The output picture collage shows as many visible salient regions (without being overlaid by others) from all images as possible. A very efficient Genetic algorithm is used here for the optimization. The experimental results showed the superior performance of the proposed method.

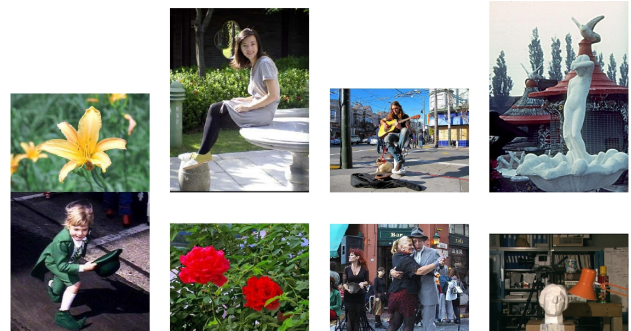

Fig. 1
LEFT IMAGES OF SOME STEREO IMAGES USED IN THIS PAPER.

## 1. Introduction

Detection of interesting or "salient" regions is a main sub-problem in the context of image tapestry and photo collage. An ideal image summary should contain as many informative regions as possible on a given space [1]. The image mosaic can be considered as a simple form of photo collage, in which the images are placed on a canvas, side by side, without any rotation or considering informative regions. Google's Picasa overlays images without considering any salient regions.

Figure 1 shows the left images of a collection of stereo images used in this paper.

Figure 2 shows some collages produced by the aforementioned methods or software over the images shown in figure 1. Images are randomly placed on a canvas. In figure 2 even the whole of all images are visible or the images are occluded, the importance of regions are discarded.

An approach named saliency-based visual attention model [2] is used in [1] for extracting interesting regions. This model combines multi scale image features (color, texture, orientation) into a single topographical saliency map. In [3] three options are considered for saliency regions:

- A heuristic assumption that the image center
Is more informative about the image's content than the border,
- it is assumed that blocks with high contrast are Salient, and
- it is assumed that typically tapestry is on a
Personal collection and a face detection module is used.

In recent years, capturing stereo images for various purposes such as 3D reconstruction has been usual. Here with this assumption that the focused object is the important part of the image and these parts are close to the camera, we used the depth map image of stereo image pairs for estimation of informative regions in a photo collage application.

The rest of the paper organized as follow: section 2 explains the proposed method. Section 3 provides the experimental results and section 4 is dedicated to the concluding remarks.

Table 1
SOME STEREO IMAGES WITH THEIR IMPORTANT REGIONS BASED ON SALIENCY MAP AND DEPTH MAP.

|  | Girl | Musician |
|---|---|---|
| Left Image | 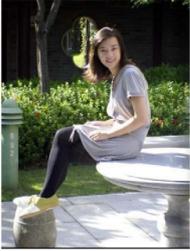 | 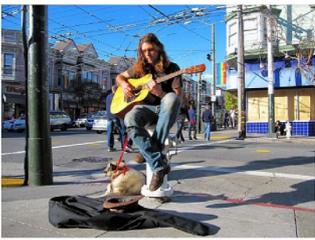 |
| Right Image | 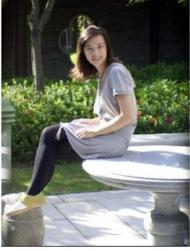 | 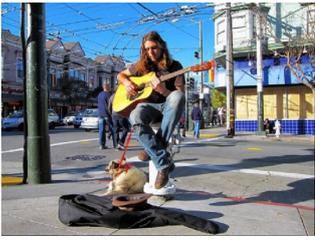 |
| Saliency Map | 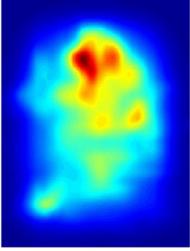 | 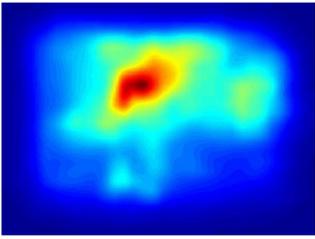 |
| Depth Map | 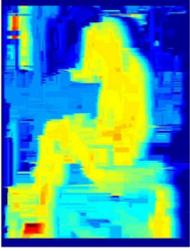 | 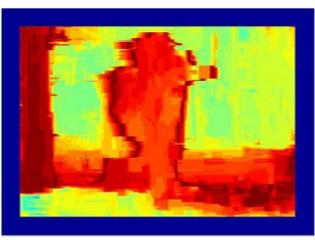 |

## 2. The Proposed Method

In order to obtain a pleasant photo collage the following properties should be considered [1], [4]:

- Salience Maximization, to maximize the total amount of visible important regions as possible ($A_{occ}$);
- Blank space minimization, to minimize the portions of the canvas which is not covered by any image ($B$),
- Salience Ratio Balance, for avoiding those cases in which a very small region of some images is visible ($V$).

The fitness function is a combination of the above three parameters: $\lambda_A.A_{occ} + \lambda_B.B + \lambda_V.V$ .

Computing the depth map has been done here with a dynamic programming approach [5]. Table 1 shows two instance image pairs and their important regions with Saliency toolbox[2][1] and stereo depth map.

Saliency map is about to left image. As can be seen, the depth map image is more informative than saliency map. The overall framework of the proposed method is shown in figure 3.

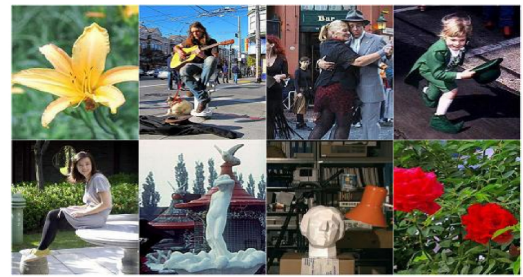

(a) Picasa Grid

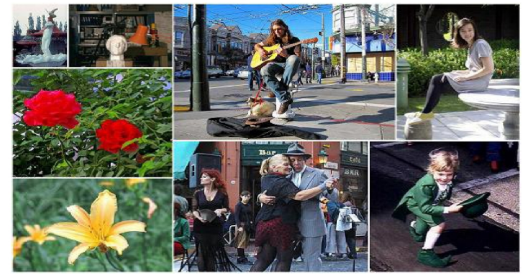

(b) Picasa Mosaic

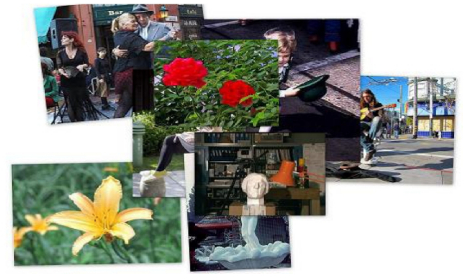

(c) Picasa Pile

Fig. 2
SOME COLLAGE OUTPUTS OF GOOGLE PICASSA.

---

[1] http://www.saliencytoolbox.net/

## 3. Experimental Results

The implementation has been done using MATLAB 7. The population size of GA was 40, number of generations was 150. In the fitness function, we set the weights $\lambda_A$, $\lambda_B$ and $\lambda_V$ as $10/30$, $9/30$ and $11/30$ respectively. The canvas is square and its size is set so that its area is about half of the total area of all input images. The input images (shown in figure 1) are gathered by searching over the Internet. Figure 4 show the value of GA fitness function over 150 generation.

Figure 5(a) shows the collage initiated at the first iteration of GA, 5(b) shows the final result of the proposed approach. Figure 5(c) shows depth map (as saliency region) occlusion, canvas usage, image rotations and cropping at the final stage of GA. As can be seen in figure 5(c) the most important regions of each image is visible in the final result.

Instead of the depth map image, we used saliency map for comparison purposes. Since the important regions with these two methods are different, the quantitative comparisons of these methods - via GA fitness function- is not possible. Hence we compared them by visual inspection of the produced collage with each method. Each method has been executed 20 times, and 2 of the best produced collages has been selected visually and illustrated in figure 6 . As can be seen the proposed method is competitive with the saliency map.

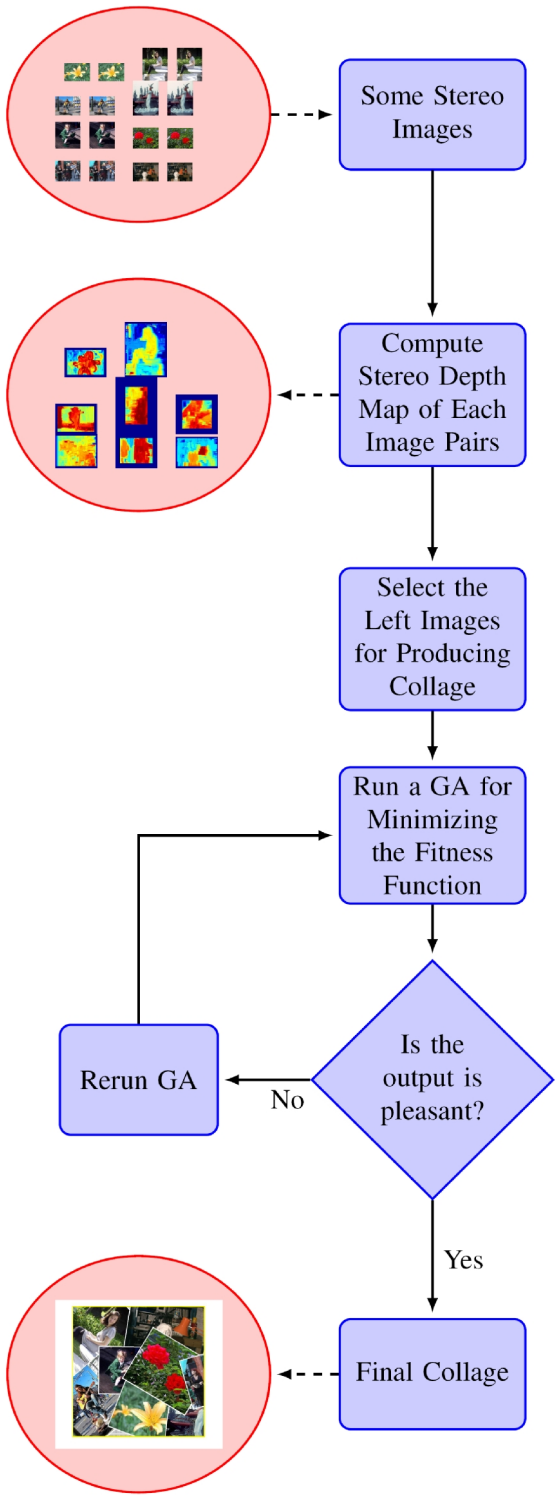

Fig. 3
THE OVERALL FRAMEWORK OF THE PROPOSED METHOD.

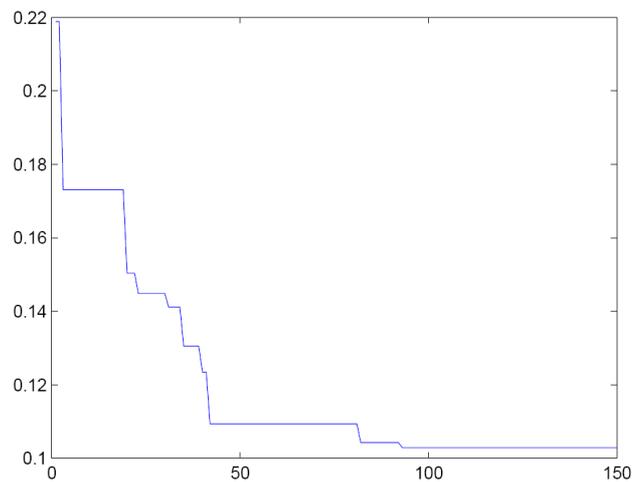

Fig. 4
THE VALUE OF FITNESS FUNCTION OVER GA ITERATIONS.

## 4. Conclusion

Until now in the photo collage context, only the mono images has been used; but here the stereo images were used for creating photo collage. In the previous works, the main source for determining important regions was based on: contrast, color, face detection and so on. Here with this assumption that usually the interested object is closer to the viewer with respect to other parts of the scene, the depth map of the stereo images has been used as an estimation of the image regions' importance.

The experimental results showed the good performance of the proposed method. Although in this paper only the stereo images has been used, but whenever some mono images and some stereo images are in hand, it is easy to use depth map for stereo images and another salient map extraction method for mono images. As future works we plan to implement the aforementioned idea and to extend our method to video collage.

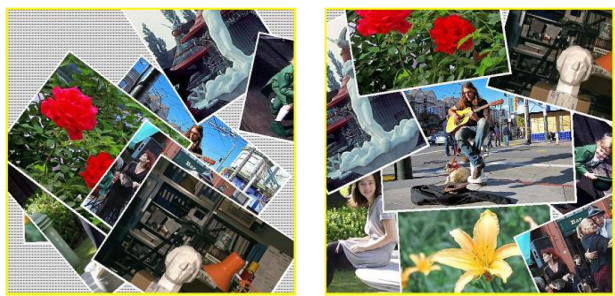

(a) First Collage at the $1^{st}$ iteration of GA.  (b) Final Result

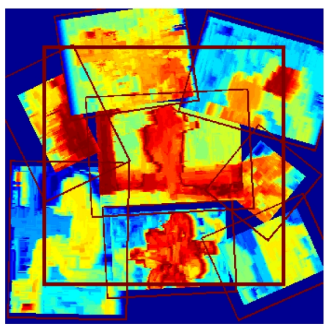

(c) Canvas Usage

Fig. 5
THE FIRST AND FINAL COLLAGE PRODUCED BY GA AT THE FIRST AND LAST ITERATIONS.

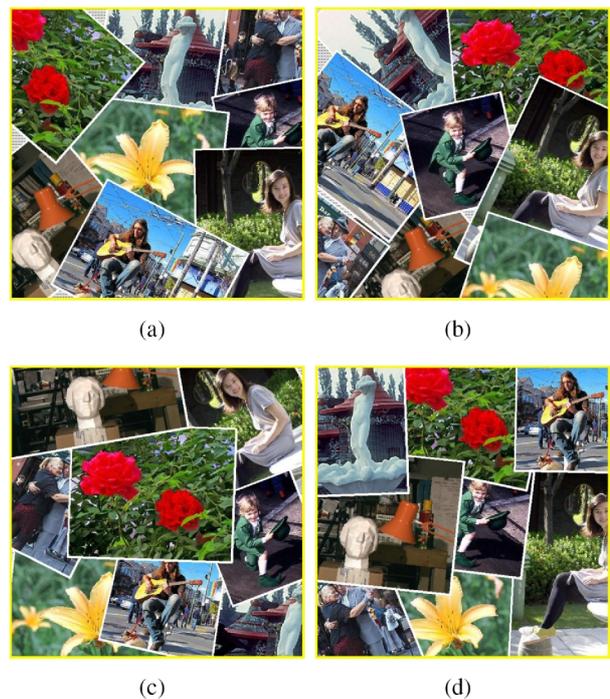

(a)  (b)

(c)  (d)

Fig. 6
THE RESULT OF THE GA, WHEN EACH OF THE SALIENCY MAP (A,B) OR DEPTH MAP (C,D) ARE USED FOR DETERMINING THE INFORMATIVE REGIONS.

**Hesam Ekhtiyar** received the B.S. degree in computer engineering from Sabzevar Tarbiat Moallem University, Sabzevar, iran, in 2011. his research interests include computer vision, speech recognition, robotics, soft computing.

**Mahdi Sheida** received the B.S. degree in computer engineering from Sabzevar Tarbiat Moallem University, Sabzevar, iran, in 2011.

his research interests include computer vision, speech recognition, network programming.

**Mahmood Amintoosi** is an assistant professor in Sabzevar Tarbiat Moallem University. He received his B.Sc. degree in Mathematics and M.Sc. degree in Computer Engineering in 1994, 1998, respectively from Ferdowsi University of Mashhad. From 1998 to 2005, he was a Lecturer in the Department of Mathematics of Sabzevar Tarbiat Moallem University.  He received his Ph.D. degree in Artificial Intelligence from Iran University of Science and Technology in 2011. His research interests include Computer Vision, Super-Resolution, Panorama, Automated Timetabling and Combinatorial Optimization. He has more than 30 conference and journal papers.